\pdfoutput=1
\documentclass[11pt]{article}

\usepackage{emnlp2021}

\usepackage{times}
\usepackage{latexsym}

\usepackage{verbatimbox} 
\usepackage{multirow}
\usepackage{makecell}
\usepackage{subcaption}
\usepackage{booktabs}
\usepackage{graphicx}
\usepackage{enumitem}
\usepackage [english]{babel}
\usepackage [autostyle, english = american]{csquotes}
\MakeOuterQuote{"}
\usepackage{subcaption}
\usepackage[T1]{fontenc}
\usepackage{amsmath}
\usepackage[utf8]{inputenc}

\usepackage{listings}
\usepackage{color}

\definecolor{dkgreen}{rgb}{0,0.6,0}
\definecolor{gray}{rgb}{0.5,0.5,0.5}
\definecolor{mauve}{rgb}{0.58,0,0.82}

\usepackage{microtype}
\usepackage{amstext, amscd, amsthm, amssymb}
\usepackage{amsmath}
\usepackage{placeins}
\usepackage{thmtools, thm-restate}
\usepackage{thm-patch, thm-kv, thm-autoref}
\usepackage{mathtools}
\usepackage{cleveref}
\declaretheorem[style=definition]{remark}

\DeclareMathOperator*{\LSE}{LSE}
\DeclareMathOperator*{\volume}{Vol}

\DeclareMathOperator*{\softplus}{softplus}

\newcommand{\BoxTensor}{\lstinline{BoxTensor}}
\newcommand{\DeltaBoxTensor}{\lstinline{MinDeltaBoxTensor}}
\newcommand{\SigmoidBoxTensor}{\lstinline{SigmoidBoxTensor}}
\newcommand{\TanhBoxTensor}{\lstinline{TanhBoxTensor}}
\DeclareMathOperator*{\logsumexp}{\mathsf{logsumexp}}
\title{Box Embeddings: An open-source library for representation learning using geometric structures}

\makeatletter
\newcommand{\printfnsymbol}[1]{%
  \textsuperscript{\@fnsymbol{#1}}%
}
\makeatother
\author{\bf Tejas Chheda\thanks{\printfnsymbol{1} Equal Contributions.} \printfnsymbol{2},
Purujit Goyal\printfnsymbol{1}\printfnsymbol{2}, Trang Tran\printfnsymbol{1}\printfnsymbol{2}\printfnsymbol{3}, Dhruvesh Patel \printfnsymbol{2},\\ {\bf Michael Boratko\printfnsymbol{2}, Shib Sankar Dasgupta\printfnsymbol{2}, \and Andrew McCallum\printfnsymbol{2}} \\
        \printfnsymbol{2} College of Information and Computer Sciences \\ 
        University of Massachusetts Amherst, MA 01003, USA \\
        \printfnsymbol{3} MassMutual  Data  Science, MA 01002, USA  \\
        \texttt{\{tchheda,purujitgoyal,ttrang,dhruveshpate\}@cs.umass.edu} \\
        \texttt{\{mboratko,ssdasgupta,mccallum\}@cs.umass.edu}}
\begin{document}
\maketitle
\begin{abstract}
\label{sec:abstract}

A major factor contributing to the success of modern representation learning is the ease of performing various vector operations.
Recently, objects with geometric structures (eg. distributions, complex or hyperbolic vectors, or regions such as cones, disks, or boxes) have been explored for their alternative inductive biases and additional representational capacities.
In this work, we introduce Box Embeddings, a Python library that enables researchers to easily apply and extend probabilistic box embeddings.
\footnote{The source code and the usage and API documentation for the library is available at \url{https://github.com/iesl/box-embeddings} and \url{https://www.iesl.cs.umass.edu/box-embeddings/main/index.html}, respectively. A quick video tutorial is available at \url{https://youtu.be/MEPDw8sIwUY}.}
Fundamental geometric operations on boxes are implemented in a numerically stable way, as are modern approaches to training boxes which mitigate gradient sparsity.
The library is fully open-source, and compatible with both PyTorch and TensorFlow, which allows existing neural network layers to be replaced with or transformed into boxes effortlessly.
In this work, we present the implementation details of the fundamental components of the library, and the concepts required to use box representations alongside existing neural network architectures.
\end{abstract}
\section{Introduction}
\label{sec:intro}

Much of the success of modern deep learning rests on the ability to learn representations of data compatible with the structure of deep architectures used for training and inference \citep{hinton2007, lecun2015}. Vectors are the most common choice of representation, as linear transformations are well understood and element-wise non-linearities offer increased representational capacity while being straightforward to implement.
Recently, various alternatives to vector representations have been explored, each with different inductive biases or capabilities.
\citet{vilnis2015word} represent words using Gaussian distributions, which can be thought of as a vector representation with an explicit parameterization of variance. This variance was demonstrated to be capable of capturing the generality of concepts, and KL-divergence provides a natural asymmetric operation between distributions, ideas which were expanded upon in \citet{athiwaratkun2018on}.
\citet{nickel2017advances}, on the other hand, change the embedding space itself from Euclidean to hyperbolic space, where the negative curvature has been shown to provide a natural inductive bias toward modeling tree-like graphs \citep{nickel2018learning, pmlr-v108-weber20a,weber2018curvature}.

A subset of these alternative approaches explores \textit{region-based} representations, where entities are not represented by a single point in space but rather explicitly parameterized regions whose volumes and intersections are easily calculated. Order embeddings \citep{vendrov2016} represent elements using infinite cones in $\mathbb R_+^n$ and demonstrate their efficacy of modeling \textit{partial orders}. \citet{lai2017learning} endow order embeddings with probabilistic semantics by integrating the space under a negative exponential measure, allowing the calculation of arbitrary marginal, joint, and conditional probabilities.
Cone representations are not particularly flexible, however - for instance, the resulting probability model cannot represent negative correlation - motivating the development of \emph{probabilistic box embeddings} \citep{vilnis2018probabilistic}, where entities are represented by $n$-dimensional rectangles (i.e. Cartesian products of intervals) in Euclidean space.

Probabilistic box embeddings have undergone several rounds of methodological improvements. The original model used a surrogate function to pull disjoint boxes together, which was improved upon in \citet{li2018smoothing} via Gaussian convolution of box indicator functions, resulting in a smoother loss landscape and better performance as a result. \citet{dasgupta2020improving} improved box training further by using a latent random variable approach, where the corners of boxes are modeled using Gumbel random variables. These latter models lacked valid probabilistic semantics, however, a fact rectified in \citet{boratko2021minmax}.

While each methodological improvement demonstrated better performance on various modeling tasks, the implementations grew more complex, bringing with it various challenges related to performance and numerical stability.
Various applications of probabilistic box embeddings (eg. modeling joint-hierarchies \citep{patel2020representing}, uncertain knowledge graph representation \citep{chen2021probabilistic}, or fine-grained entity typing \citep{onoe2021modeling}) have relied on bespoke implementations, adding unnecessary difficulty and differences in implementation when applying box embeddings to new tasks.
To mitigate this issue and make applying and extending box embeddings easier, we saw the need to introduce a reusable, unified, stable library that provides the basic functionalities needed in studying box embeddings. To this end, we introduce ``Box Embeddings'', a fully open-source Python library hosted on PyPI. 
The contributions of this work are as follows:
\begin{itemize}[noitemsep,topsep=0pt]
    \item Provide a modular and reusable library that aids the researchers in studying probabilistic box embeddings. The library is compatible with both of the most popular Machine Learning libraries: PyTorch and TensorFlow.
    \item Create extensive documentation and example code, demonstrating the use of the library to make it easy to adapt to existing code-bases.
    \item Rigorously unit-test the codebase with high coverage, ensuring an additional layer of reliability.
\end{itemize}

\section{Box Embeddings}
\label{sec:design and implementation}
Formally, a "box" is defined as a Cartesian product of closed intervals,
\begin{align*}
    B(\theta) &= \prod_{i=1}^n [z_i(\theta), Z_i(\theta)]\\
    &= [z_1(\theta), Z_1(\theta)] \times \cdots \times [z_n(\theta), Z_n(\theta)],
\end{align*}
where $\theta$ represent some latent parameters. In the simplest case, $\theta \in \mathbb R^{2n}$ are free parameters, and $z_i, Z_i$ are projections onto the $i$ and $n+i$ components, respectively. In general, however, the parameterization may be more complicated, eg. $\theta$ may be the output from a neural network.
For brevity, we omit the explicit dependency on $\theta$.
The different operations (such as volume and intersection) commonly used when calculating probabilities from box embeddings can all be defined in terms of $z_i, Z_i$ - the \emph{min} and \emph{max} coordinates of the interval in each dimension.

\subsection{Parameterizations}
\label{sec:parameterizations}
\begin{figure}[h!]
    \centering
    \includegraphics[width=0.9\linewidth]{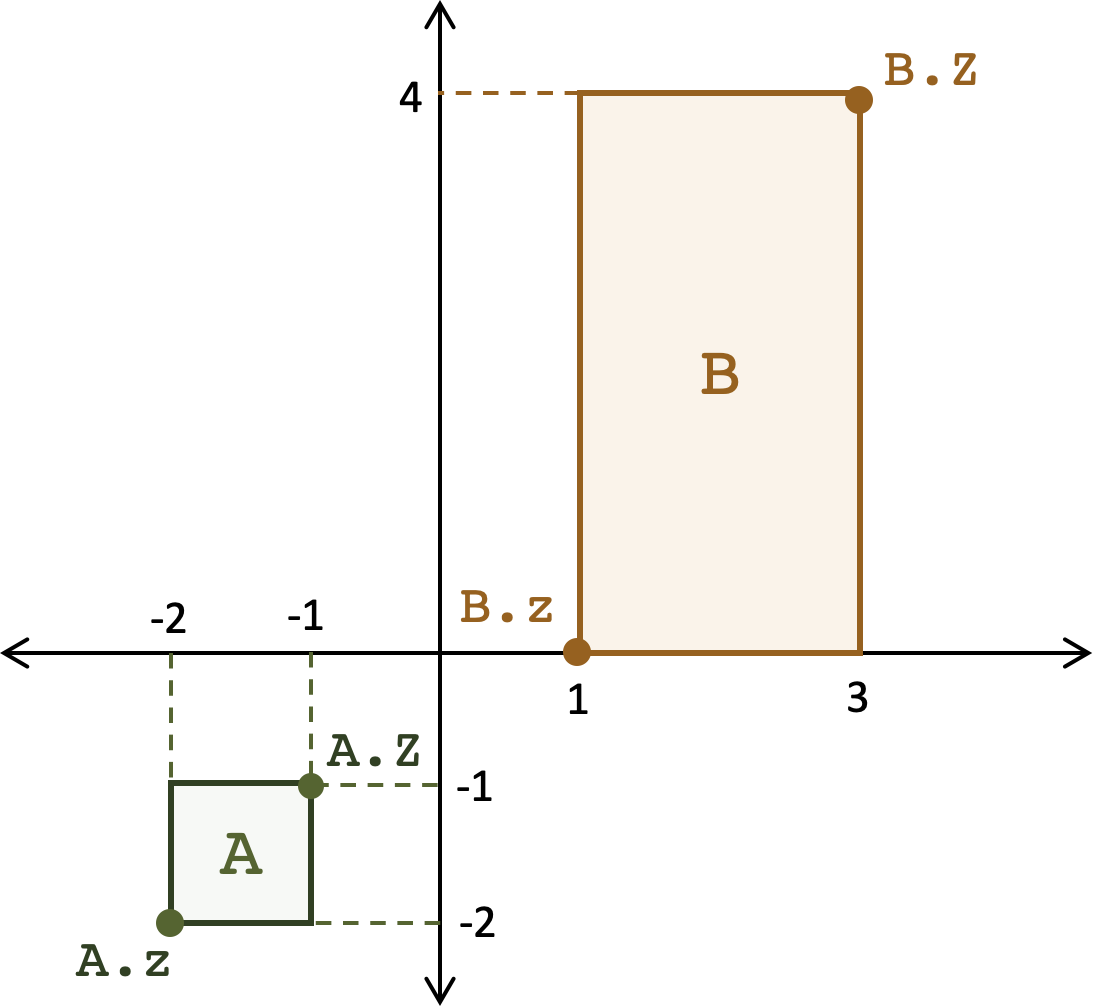}
\caption{Box Parameterization}
\label{fig: box parameterization}
\end{figure}

The fundamental component of the library is the \BoxTensor{} class, a wrapper around the \lstinline{torch.Tensor} and \lstinline{tensorflow.Tensor} class that represents a tensor/array of boxes. 
\BoxTensor{} is an opaque wrapper, in that it exposes the operations and properties necessary to use the box representations (see table \ref{tab:box tensor methods}) irrespective of the specific way in which the parameters $\theta$ are related to $z_i, Z_i$.
The main two properties of the \BoxTensor{} are \lstinline{z} and \lstinline{Z}, which represent the \emph{min} and \emph{max} coordinates of an instance of \BoxTensor{}. Listing \ref{lst:init} shows how to create an instance of \BoxTensor{} consisting of two 2-dimensional boxes in Figure \ref{fig: box parameterization}.

\begin{table}
\centering
\resizebox{0.48\textwidth}{!}{%
\begin{tabular}{|l|l|}
\hline
$z$                         & the lower-left coordinate of the boxes                   \\ \hline
$Z$                         & the top-right coordinate of the boxes                    \\ \hline
\lstinline{centre}          & the center coordinate of the boxes, $\frac{z + Z}{2}$    \\ \hline
\lstinline{box\_shape}      & shape of the center coordinates (or $z, Z$)                 \\ \hline
\lstinline{box\_reshape}    & if possible, reshapes the \lstinline{box\_shape} into the \lstinline{target\_shape} \\ \hline
\lstinline{broadcast}       & \makecell[l]{if possible, adds new dimensions to the \lstinline{box\_shape} to make \\ it compatible with the \lstinline{target\_shape}} \\ \hline
\end{tabular}
}
\caption{BoxTensor Properties}
\label{tab:box tensor methods}
\end{table}

\begin{lstlisting}[caption={Manually initializing a \lstinline{BoxTensor} consisiting for the 2-D boxes depicted in \Cref{fig: box parameterization}.},label={lst:init}]
    import torch
    from box_embeddings.parameterizations import BoxTensor
    theta = torch.tensor(
        [[[-2, -2], [-1, -1]], [[1, 0], [3, 4]]]
    )
    box = BoxTensor(theta)
    A = box[0]
    B = box[1]
\end{lstlisting}

Given a \lstinline{torch.Tensor} corresponding to the parameters $\theta$ of a \BoxTensor{}, one can obtain a box representation in multiple ways depending on the constraints on the \emph{min} and \emph{max} coordinates of the box representations as well as the the range of values in $\theta$.
The \BoxTensor{} class itself simply splits $\theta$ in half on the last dimension, using $\theta[\dots, 1:n]$ as $z$ and $\theta[\dots, n+1:2n]$ as $Z$. Here, the \lstinline{Ellipsis} "$\dots$" denotes any number of leading dimensions, for instance, batch, sequence-length, etc. For the sake of simplifying the notations, from here on, the presence of the leading dimensions will not be explicitly denoted using the \lstinline{Ellipsis}. Moreover, all the indexing operations can be assumed to be operating only on the last dimension, unless stated otherwise. 

\begin{lstlisting}[caption={Converting latent vectors to boxes, for various choices of box parameterizations.},label={lst:params}]
from box_embeddings.parameterizations import BoxTensor, MinDeltaBoxTensor, SigmoidBoxTensor
box_tensor = BoxTensor(theta)
box_tensor_pos_sides = MinDeltaBoxTensor(theta)
box_tensor_in_unit_cube = SigmoidBoxTensor(theta)
\end{lstlisting}

Any box can be represented in this fashion, however some settings of $\theta$ may lead to situations where $z_i > Z_i$.
This scenario is invalid under conventional box models \citep{vilnis2018probabilistic, li2018smoothing}, and although valid for models which interpret these coordiantes as parameters of a latent random variable \citep{dasgupta2020improving, boratko2021minmax} it is often still desirable to constrain side-lengths to be non-negative.
\lstinline{MinDeltaBoxTensor} represents boxes that are unbounded and have non-negative side-length in each dimension. That is, it outputs boxes with $z,Z \in \mathbb R^{n}$ and $z_i\le Z_i$, and furthermore any such box has a corresponding $\theta$ under this parameterization.
A valid probabilistic interpretation of box embeddings requires that their embedding space has finite measure, however.
One trivial way to accomplish this is to parameterize boxes to remain within the unit hypercube, which can be accomplished via the
\lstinline{SigmoidBoxTensor} or \lstinline{TanhBoxTensor} classes.
The specific mathematical operations relating the $\theta$ variables to their $z, Z$ coordinates are found in \Cref{tab:param}, and example usage can be found in \Cref{lst:params}.\footnote{The TensorFlow version for all the code snippets is provided in Appendix.}

\begin{table}
\centering
\resizebox{0.48\textwidth}{!}{%
\begin{tabular}{@{}lll@{}}
\toprule
Parameterization            & $z$           & $Z$              \\ \midrule
\BoxTensor{}                   & $\theta[1:n]$ & $\theta[n+1:2n]$ \\
\DeltaBoxTensor{} & $\theta[1:n]$ & $z+\softplus(\theta[n+1:2n])$ \\
\SigmoidBoxTensor & $\sigma(\theta[1:n])$ & $ z+(1-z)\sigma(\theta[n+1:2n])$\\
\TanhBoxTensor    & $\frac{\tanh(\theta[1:n])+1}{2}$            & $z + \frac{(1-z)\tanh(\theta[n+1:2n])}{2}$               \\ \bottomrule
\end{tabular}
}
\caption{The different subclasses of \BoxTensor{} and how they represent boxes using the learnable parameters $\theta\in \mathbb R^{2n}$ taken as input.}
\label{tab:param}
\end{table}

\subsection{Operations on BoxTensor}
\label{sec:operations}

We provide a variety of modules that implement different operations on the box-tensors, such as \lstinline{Intersection}, \lstinline{Volume}, \lstinline{Pooling} and \lstinline{Regularization}. We also implemented a \lstinline{BoxEmbedding} layer that, just like a vector embedding layer, provides index lookup. However, unlike a vector embedding layer, this returns boxes instead of vectors. 
We discuss these layers in detail below.

\subsubsection{Intersection}

Given two instances of \BoxTensor{} with compatible shapes, this operation performs the intersection between the two box-tensors and returns an instance of \BoxTensor{} as the result. For two instances of \BoxTensor{} $A$ and $B$ with coordinates $(z_A,Z_A)$ and $(z_B,Z_B)$ respectively, the $(z, Z)$ coordinates of the resulting intersection box for the two types of intersection operations, \lstinline{HardIntersection} \cite{vilnis2018probabilistic,li2018smoothing} and \lstinline{GumbelIntersection} \cite{dasgupta2020improving}, are shown in \Cref{tab:intersection}, and corresponding codes are provided in \Cref{lst:intersection}.

\begin{lstlisting}[caption={Various approaches to computing the intersection of two box tensors.},label={lst:intersection}]
from box_embeddings.parameterizations import BoxTensor
from box_embeddings.modules.intersection import HardIntersection, GumbelIntersection

boxA = BoxTensor(theta_a)
boxB = BoxTensor(theta_b)

hard_intersection = HardIntersection()
gumbel_intersection = GumbelIntersection(intersection_temperature=0.8)

hard_ab = hard_intersection(boxA, boxB)
gumbel_ab = gumbel_intersection(boxA, boxB)
\end{lstlisting}

\begin{table}
\centering
\resizebox{0.48\textwidth}{!}{%
\begin{tabular}{@{}lll@{}}
\toprule
Intersection type  & $z$                                               & $Z$                                                  \\ \midrule
\lstinline{HardIntersection}   & $\max(z_A, z_B)$                                  & $\min(Z_A, Z_B)$                                     \\
\lstinline{GumbelIntersection} & $\beta \LSE(\frac{z_A}{\beta},\frac{z_B}{\beta})$ & $-\beta \LSE(-\frac{Z_A}{\beta},-\frac{Z_B}{\beta})$\\\bottomrule
% MB: I changed lae to LSE (not sure if people know what lae means)
\end{tabular}%
}
\caption{Expressions for the two kinds of intersection layers. Here, $\LSE$ denotes $\logsumexp$, i.e., $\LSE(x,y):=\log(\exp(x) + \exp(y))$}
\label{tab:intersection}
\end{table}

\subsubsection{Volume}
\label{sec:volume}
Boxes (or intersections of boxes) are typically queried for their \textit{volumes}. Our \lstinline{HardVolume} layer implements the volume calculation as originally introduced in \citet{vilnis2018probabilistic}, which is simply a direct multiplication of side-lengths. It is in this setting where bounded parameterizations such as \SigmoidBoxTensor{} and \TanhBoxTensor{} are particularly useful, as the resulting volumes can be interpreted as yielding a valid marginal or joint probability. Note, however, that the guarantees of positive side-lengths do not apply when taking the intersection of two disjoint boxes, in which case the resulting box should have zero volume.

Our \lstinline{SoftVolume} layer implements the volume function proposed by \citet{li2018smoothing}, which mitigates the training difficulties that arise when disjoint boxes should overlap. Finally, our \lstinline{BesselApproxVolume} layer implements the volume function proposed in \citet{dasgupta2020improving}, which approximates the expected volume of a box where the coordinates are interpreted as location parameters of Gumbel random variables. The expressions and the code snippets for the various volume operations are given in Table \ref{tab:volume} and \ref{lst:volume}, respectively.

\begin{remark}
Note that due to the presence of the product, the naive implementation of volume computations as shown in Table \ref{tab:volume} will often result in numerical overflow or underflow for dimensions greater than 5.
Hence, we provide an option to compute the volume in log-space, which is \emph{on} by default.
\end{remark}

\begin{lstlisting}[caption={Different proposed methods for computing box volume, of increasing "smoothness".},label={lst:volume}]
from box_embeddings.modules.volume import HardVolume, SoftVolume, BesselApproxVolume

hard_volume = HardVolume()
log_volA = hard_volume(boxA)

soft_volume = SoftVolume(volume_temperature=5.0)
log_vol_ab = soft_volume(hard_ab)

bessel_volume = BesselApproxVolume(volume_temperature=5.0, intersection_temperature=0.8)
log_vol_ab = bessel_volume(gumbel_ab)
\end{lstlisting}

\begin{table}
\centering
\resizebox{0.48\textwidth}{!}{%
\begin{tabular}{@{}ll@{}}
\toprule
Intersection type & Volume                       \\ \midrule
\lstinline{HardVolume}        & $\prod_{i=1}^n \max(Z_i - z_i, 0) $ \\
\lstinline{SoftVolume}         & $\prod_{i=1}^n T*\softplus( \frac{Z_i - z_i}{T})$         \\
\lstinline{BesselApproxVolume} & $\prod_{i=1}^n T*\softplus(\frac{Z_i - z_i -2\gamma \beta}{T})$ \\ \bottomrule
\end{tabular}%
}
\caption{The expressions for different volume implementations. Here, $(z,Z)$ are the min-max coordinates of the input \BoxTensor, $T$ is the volume temperature hyperparameter, $\gamma$ is the Euler-Mascheroni constant, $\beta$ is the gumbel intersection parameter, and $\softplus(x)=\log(1+\exp x)$.}
\label{tab:volume}
\end{table}

\subsubsection{Pooling}
\label{sec:pooling}
The library also provides pooling operations that take as input an instance of \BoxTensor{} and reduce one of the leading dimensions by pooling across it. 
Currently, there are two types of pooling operations implemented -- intersection based, which takes intersection across all the boxes in a particular dimension, and mean based, which takes the arithmetic mean of the min and max coordinates of the boxes across a dimension. 

\subsubsection{Regularization}
\label{sec:regularization}
There is an excessive slackness in the learning objective defined using containment conditions on boxes, which leads to large flat regions of local minima resulting in poor training. 
In order to mitigate this problem, \citet{patel2020representing} introduces volume based regularization for boxes, which augments the loss with a penalty if the box volume exceeds a certain threshold. 
This penalty reduces the size of the flat local minima facilitating better training of boxes.

\begin{lstlisting}[caption={Box pooling and regularization operations.},label={lst:pool_n_reg}]
from box_embeddings.modules.pooling import HardIntersectionBoxPooler
from box_embeddings.modules.regularization import L2SideBoxRegularizer

pooler = HardIntersectionBoxPooler()
pooled_box = pooler(box)

box_regularizer = L2SideBoxRegularizer(log_scale=True)

vol_box = soft_volume(pooled_box)
loss = loss_fn(vol_box) + box_regularizer(pooled_box)
\end{lstlisting}

\subsection{Embedding}
\label{sec:embedding}
\BoxTensor{} and its children classes, do not store learnable parameters directly, they simply wrap the input tensor and provide an interface which interprets the wrapped tensor as box representation.
However, when working with a shallow model (embedding only model), one needs an embedding layer that owns its parameters and outputs boxes corresponding to the input indices.
The library provides \lstinline{BoxEmbedding} layer that works like a native embedding layer in PyTorch or TensorFlow, i.e., it performs index lookup, but instead of returning an instance of the native tensor, it returns instance of \BoxTensor{}.

\subsubsection{Initializers}
\label{sec:init}

We also provide an abstract interface \lstinline{BoxInitializer} to implement various methods for initializing the learnable parameters of the \lstinline{BoxEmbedding} layer. 
As a concrete example we implement \lstinline{UniformBoxInitializer}, which initializes boxes with uniformly random min coordinates and side lengths. This is used as the default initializer for the \lstinline{BoxEmbedding} layer unless specified otherwise.

\section{Applications}
\label{sec: applications}

\begin{figure*}[htp!]
\centering
\subcaptionbox{An example hierarchical structure\label{fig:hierarchical_graph}}
[0.45\linewidth]{\includegraphics[width=0.9\linewidth]{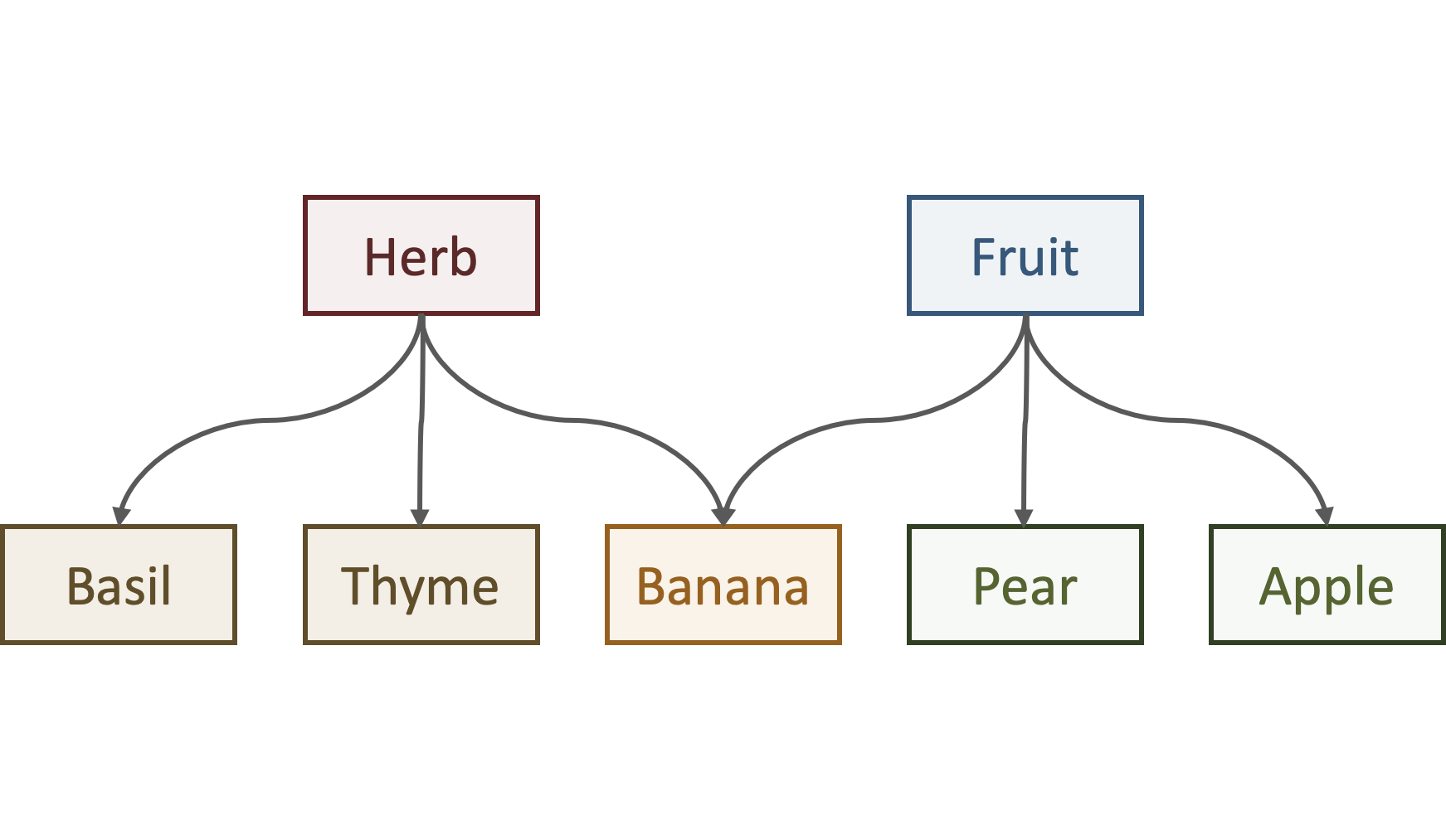}}
\subcaptionbox{Representing the structure in (a) with Box Embeddings\label{fig:hierarchy_boxes}}
[0.45\linewidth]{\includegraphics[width=0.7\linewidth]{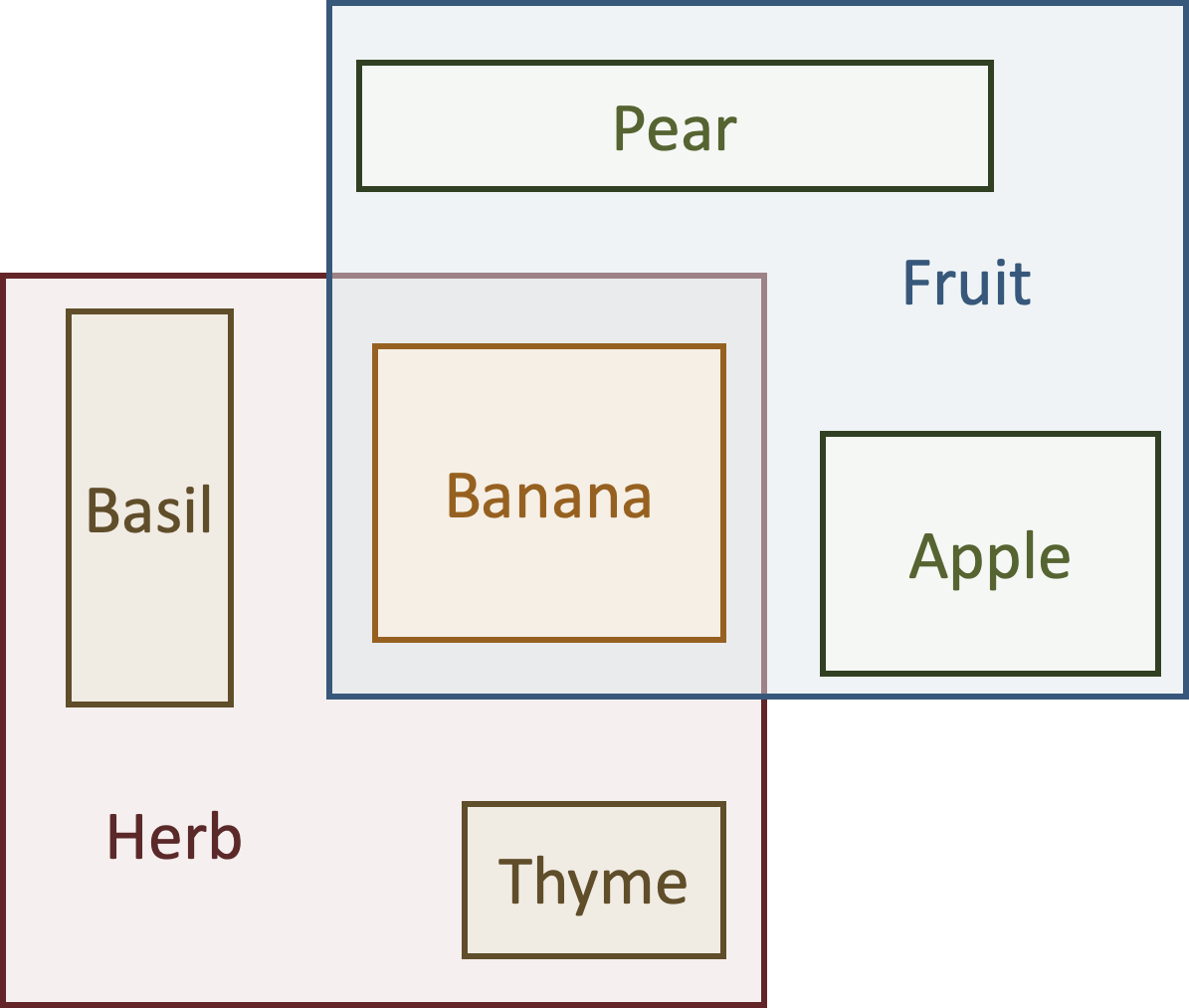}}

\caption{Box Embeddings can capture hierarchical structures commonly observed in natural language}
\label{fig:application}
\end{figure*}
In this section, we demonstrate the Box Embeddings library by using it to implement models for two real-world tasks: a representation learning task of hierarchical graph modeling \cite{nickel2017advances,vilnis2018probabilistic}, and the NLP task of natural language inference \cite{dagan2005pascal,bowman2015large}. We first demonstrate the intuition behind the containment-based loss function used to train these models using a toy example involving two 2-dimensional boxes.

\subsection{Toy example} \label{subsec:toy}
\begin{figure}[htp!]
\centering
\begin{subfigure}{.5\linewidth}
  \centering
  \includegraphics[width=\linewidth]{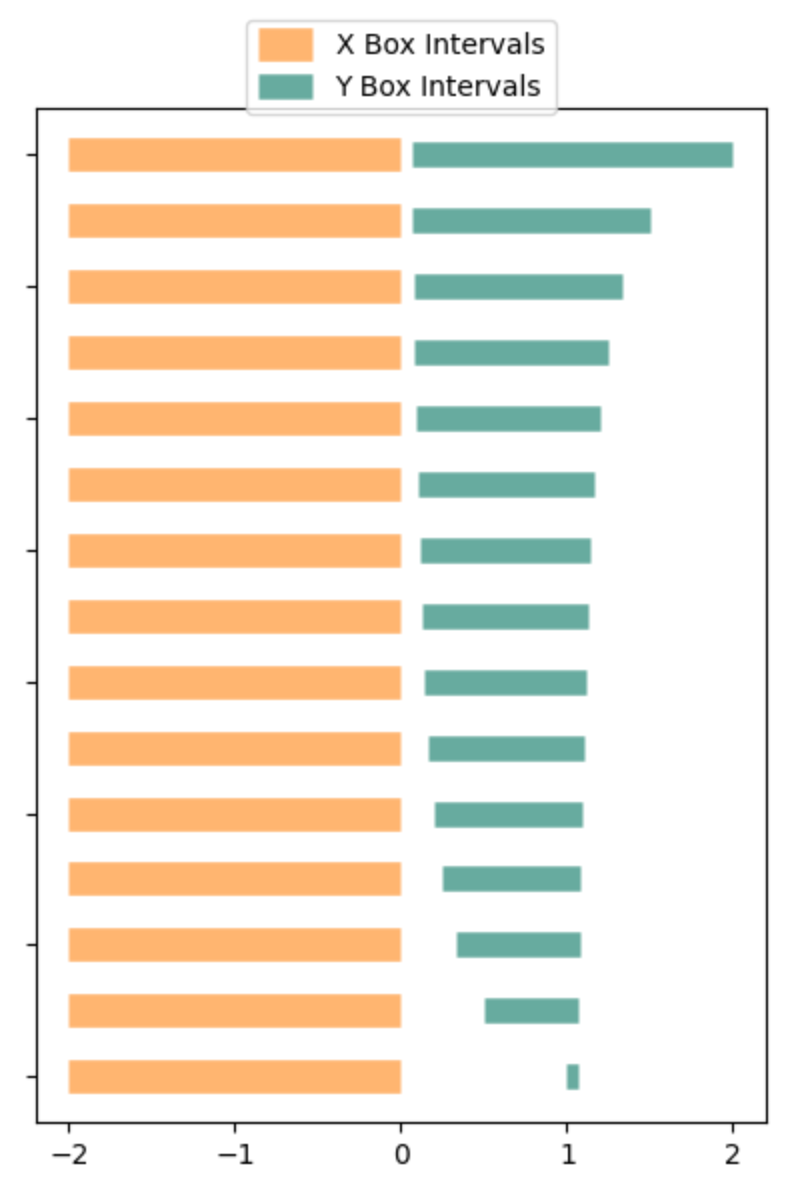}
  \caption{Before training}
  \label{fig:visualization_before}
\end{subfigure}%
\begin{subfigure}{.5\linewidth}
  \centering
  \includegraphics[width=0.996\linewidth]{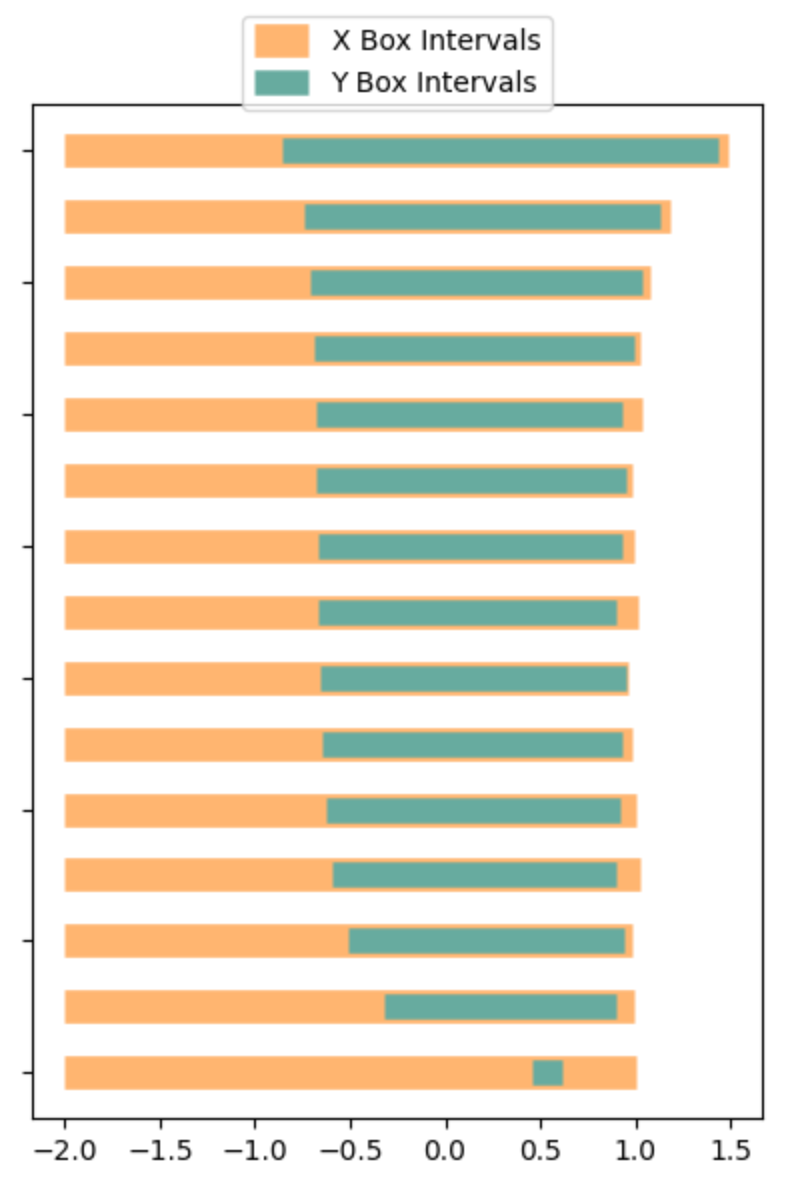}
  \caption{After training}
  \label{fig:visualization_after}
\end{subfigure}
\caption{Visualization of two 15-dimensional boxes before and after containment training described in Section \ref{subsec:toy}. The green box $B(\theta_Y)$ has been trained to be entirely contained in the orange box $B(\theta_X)$.}
\label{fig:containment}
\end{figure}

For the purpose of demonstration, we set up a toy example which embeds a simple graph with just two nodes, $X, Y$ and one edge $(X, Y)$. We start with two non-overlapping boxes at initialization: box$_X$ and box$_Y$, and use SGD to train the parameters that minimize the following loss function
$$
\mathcal{L}(\theta) = -\log \frac
{\volume{(\text{B}(\theta_X) \cap \text{B}(\theta_Y)})}
{\volume{(\text{B}(\theta_Y))}}
.   
$$
Geometrically, this encourages $\text{box}_Y \subseteq \text{box}_X$. If using a box embedding with valid probabilistic semantics, this loss function can be interpreted as binary cross-entropy with $P(X | Y) = 1$.\footnote{To understand further the motivation for this choice of graph embedding, see \citet{vilnis2018probabilistic}.}
The code for this example can be found in Appendix \ref{appendix:example}. We visualize the containment training process in Figure \ref{fig:containment}. 
Each line represents the edge of the box in one dimension, with the left endpoint of a blue or orange line to be the minimum coordinate of a box, and the right endpoint of a line to be the maximum coordinate of a box.

\subsection{Representing hierarchical graph}

Representing relations between the nodes of a hierarchy is useful for various NLP and Machine Learning tasks such as natural language inference \cite{Wang_Kapanipathi_Musa_Yu_Talamadupula_Abdelaziz_Chang_Fokoue_Makni_Mattei_Witbrock_2019,sharma-etal-2019-incorporating}, entity typing \cite{onoe2021modeling}, multi-label classification \cite{chatterjee-etal-2021-joint}, and question answering \cite{jin2019comqa,fang2020hierarchical}.
For example, in Figure \ref{fig:application}, knowing the hypernym relationship between the pairs \textit{(herb, basil)}, \textit{(herb, thyme)}, and \textit{(herb, rosemary)} can help paraphrase the sentence ``This dish requires basil, thyme and rosemary'' into ``This dish requires several herbs.''. Additionally, knowing the relationship between \textit{(herb, banana)}, and \textit{(fruit, banana)} can help answer questions such as "What is both a herb and a fruit?" 
Note that this latter example maps directly onto the notion of box intersection, as we are seeking an element contained in both "herb" and "fruit".

For demonstration, we train box embeddings to represent the hypernym graph of WordNet \cite{miller1990introduction}. 
Hypernym or IS-A is a transitive relation between a pair of words, where one word (hypernym) represents a general/broader concept, and the other word (hyponym) is a more specific sub-concept \citep{yu2015learning}.
The transitive reduction of the WordNet noun hierarchy contains 82,114 entities and 84,363 edges. 
The learning task is framed as an edge classification task where, given a pair of nodes $(h,t)$, the model outputs the probability of existence of an edge from $h$ to $t$.
Following \citet{patel2020representing}, we train an edge classification model using the transitive reduction edges augmented with varying percentages of the transitive closure edges (10\%, 25\%, 50\%) as positive examples and randomly sampled negative examples with positive to negative ratio of 1:10. 
The \lstinline{BoxEmbedding} layer is initialized with random boxes representing the nodes of the hypernym graph.
%, box$_H$ and box$_T$ respectively. 
For each input pair $x=(h_i, t_i)$, the probability of existence of the edge $h_i\to t_i$ is computed as 
\begin{align*}
    P(h_i\to t_i) = \frac{\volume{(\text{B}(\theta_{h_i}) \cap \text{B}(\theta_{t_i}))}}{\volume{(\text{B}(\theta_{t_i}))}}.
\end{align*}
In our case, we use \DeltaBoxTensor{} parameterization, \lstinline{HardIntersection} and \lstinline{SoftVolume}. 
Binary cross-entropy loss is used to train the model for edge classification.
The test set consists of positive edges sampled from the rest of the transitive closure (not seen during training) and a fixed set of random negatives with the same positive to negative ratio as training.  As seen in Table \ref{tab:wordnet}, we are able to replicate the result from \citet{patel2020representing}.

\begin{table}
\centering
\resizebox{0.48\textwidth}{!}{%
\begin{tabular}{l | cccc}
\toprule
\multicolumn{1}{c}{TC Edges}            & 0\%    & 10\%   & 25\%   & 50\%   \\ \midrule
w/o Regularization & 44.2\% & 71.3\% & 81.1\% & 89.1\% \\
w Regularization   & 59.4\% & 90.3\% & 91.9\% & 94.2\% \\ \bottomrule
\end{tabular}
}
\caption{Test F1 scores for predicting the transitive closure of WordNet’s hypernym relations when training on increasing amounts of edges from the transitive closure}
\label{tab:wordnet}
\end{table}

\subsection{Natural Language Inference (NLI)}
\label{sec:nli}
Natural language inference \cite{dagan2005pascal,bowman2015large} is a task where, given two sentences, premise and hypothesis, the model is required to pick whether the premise entails the hypothesis, contradicts the hypothesis, or whether neither relationship holds.
The task of NLI is setup as multi-class classification, and in the two-class version, the model is only required to decide whether the premise entails the hypothesis or not \cite{mishra-etal-2021-looking}.
Although NLI deals with a pair of sentences at a time, in the space of all possible sentences the transitive relation of entailment establishes a partial order. 
If the sentences are encoded as boxes then we can train box containment to capture the transitive entailment relation.
To demonstrate this, we choose the MNLI corpus \cite{N18-1101} from the GLUE benchmark \cite{wang-etal-2018-glue}. 
Since the MNLI dataset presents the NLI task as a three-class problem, we collapse \emph{contradiction} and \emph{neutral} labels into a single label called \emph{not-entails} to obtain a two-class problem with class labels \emph{entails} and \emph{not-entails}. 

In order to obtain box representation for the premise and hypothesis sentences, we use a neural network $E$ to first get vector representations $v_p$ and $v_h$ for the premise and the hypothesis, respectively. 
Both these vectors are then interpreted as the parameters $\theta_p:=v_p$ and $\theta_h:=v_h$ of a box tensor.
Finally, the probability of the \emph{entails} class is computed as 
$$
    P(\text{entails}) = \frac{\volume{(\text{B}(\theta_p) \cap \text{B}(\theta_h))}}{\volume{(\text{B}(\theta_h))}}.
$$
The parameters of the encoder are trained using the ADAM optimizer \cite{kingma2014adam} with binary cross-entropy as the loss. 
Table \ref{tab:mnli} shows the test accuracy with two different encoders. As seen, the performance is much higher than random or majority class baselines. 

\begin{table}
\centering
\begin{tabular}{ll}
\toprule
\multicolumn{1}{c}{Neural Network Encoder (E)}                                         & Accuracy                  \\ \midrule
\multicolumn{1}{l}{RoBERTa} & \multicolumn{1}{c}{78\%} \\
\multicolumn{1}{l}{LSTM}    & \multicolumn{1}{c}{73\%} \\ \midrule
\multicolumn{1}{l}{Random Baseline}    & \multicolumn{1}{c}{50\%} \\ 
\multicolumn{1}{l}{Majority Baseline}    & \multicolumn{1}{c}{66\%} \\ \bottomrule
\end{tabular}
\caption{Test accuracy on MNLI task using box embeddings}
\label{tab:mnli}
\end{table}

\section{Conclusion}
In this paper, we have introduced Box Embeddings, the first Python library focused on allowing region-based representations to be used with deep learning libraries. Our library implements proposed training methods and geometric operations on probabilistic box embeddings in a well-tested and numerically-stable fashion. We described the concepts needed to understand and apply this library to novel tasks, and applied the library to graph modeling and natural language inference, demonstrating both shallow and deep contextualized box representations. We hope the release of this package will aid researchers in using region-based representations in their work, and that the well-documented codebase will facilitate additional methodological extensions to probabilistic box embedding models.

\section*{Acknowledgements}
The authors would like to thank the members of the Information Extraction and Synthesis Laboratory (IESL) at the University of Massachusetts Amherst for their tremendous support and feedback throughout the project. We also thank Lorraine Li for her involvement in the early phase of the project. This work is funded in part by the University of Southern California subcontract no. 123875727 under Office of Naval Research prime contract no. N660011924032 and University of Southern California subcontract no. 89341790 under Defense Advanced Research Projects Agency prime contract no. FA8750-17-C-0106.
The U.S. Government is authorized to reproduce and distribute reprints for Governmental purposes notwithstanding any copyright notation thereon. The views and conclusions contained herein are those of the authors and should not be interpreted as necessarily representing the official policies or endorsements, either expressed or implied, of DARPA or the U.S. Government.
\bibliography{anthology,custom}
\bibliographystyle{acl_natbib}
\newpage
% removing clearpage and newpage because the camera-ready instructions ask us to not start appendices on a new page.
%\clearpage
%\newpage
\appendix
\section{Appendix}
\label{sec:appendix}
\subsection{TensorFlow (TF) version}

\begin{lstlisting}[caption={TF code for initializing a \lstinline{BoxTensor}.}]
    import tensorflow as tf
    from box_embeddings.parameterizations import TFBoxTensor
    theta = tf.Variable(
        [[[0, 0], [2, 2]], [[4, 0], [8, 4]]]
    )
    box = BoxTensor(theta)
    boxA = box[0]
    boxB = box[1]
\end{lstlisting}

\begin{lstlisting}[caption={TF code for converting theta vectors to boxes.}]
from box_embeddings.parameterizations import TFMinDeltaBoxTensor, TFSigmoidBoxTensor, TFTanhBoxTensor 
box_tensor = TFMinDeltaBoxTensor(theta)
box_tensor_pos_sides = TFSigmoidBoxTensor(theta)
box_tensor_in_unit_cube = TFTanhBoxTensor(theta)
\end{lstlisting}

\begin{lstlisting}[caption={TF code for computing the intersection of two box tensors.}]
from box_embeddings.parameterizations import TFBoxTensor
from box_embeddings.modules.intersection import TFHardIntersection
from box_embeddings.modules.intersection import TFGumbelIntersection

boxA = TFBoxTensor(theta_a)
boxB = TFBoxTensor(theta_b)

hard_intersection = TFHardIntersection()
gumbel_intersection = TFGumbelIntersection()

hard_ab = hard_intersection(boxA, boxB)
gumbel_ab = gumbel_intersection(boxA, boxB)
\end{lstlisting}

\begin{lstlisting}[caption={TF code for computing the volume of a box.}]
from box_embeddings.modules.volume import TFHardVolume
from box_embeddings.modules.volume import TFSoftVolume
from box_embeddings.modules.volume import TFBesselApproxVolume

hard_volume = TFHardVolume()
volA = hard_volume(boxA)

soft_volume = TFSoftVolume()
vol_ab = soft_volume(hard_ab)

bessel_volume = TFBesselApproxVolume()
vol_ab = bessel_volume(gumbel_ab)
\end{lstlisting}

\begin{lstlisting}[caption={TF code for performing pooling and regularization operations over a box.}]
from box_embeddings.modules.pooling import TFHardIntersectionBoxPooler
from box_embeddings.modules.regularization import TFL2SideBoxRegularizer

pooler = TFHardIntersectionBoxPooler()
pooled_box = pooler(box)

box_regularizer = TFL2SideBoxRegularizer(log_scale=True)

vol_box = soft_volume(pooled_box)
loss = loss_fn(vol_box) + box_regularizer(pooled_box)
\end{lstlisting}

\subsection{Toy Example} \label{appendix:example}
\begin{lstlisting}[caption={Training Pipeline for the Toy Example (\ref{subsec:toy})}]
import torch
import numpy
from box_embeddings.parameterizations.box_tensor import BoxTensor
from box_embeddings.modules.volume.volume import Volume
from box_embeddings.modules.intersection import Intersection

# Initialization
x_z = numpy.array([-2.0 for n in range(1, 16)])
x_Z = numpy.array([0.0 for k in (x_z)])
data_x = torch.tensor([x_z, x_Z], requires_grad=True)
box_H = BoxTensor(data_x)

y_z = numpy.array([1/n for n in range(1, 16)])
y_Z = numpy.array([1 + k for k in reversed(y_z)])
data_y = torch.tensor([y_z, y_Z], requires_grad=True)
box_T = BoxTensor(data_y)

# Training function
learning_rate = 0.1
def train(box_1, box_2, optimizer, epochs=1):
    best_loss = int()
    best_box_1 = None
    best_box_2 = None
    box_vol = Volume(volume_temperature=0.1, intersection_temperature=0.0001)
    box_int = Intersection(intersection_temperature=0.0001)
    for e in range(epochs):
        loss = box_vol(box_2) - box_vol(box_int(box_1, box_2))
        optimizer.zero_grad()
        loss.backward()
        optimizer.step()
        if best_loss < loss.item():
            best_loss = loss.item()
            best_box_2 = box_2
            best_box_1 = box_1
        print('Iteration %d, loss = %.4f' % (e, loss.item()))
    return best_box_1, best_box_2

# Train
optimizer =  torch.optim.SGD([data_x, data_y], lr=learning_rate)
best_box_H, best_box_T = train(box_H, box_T, optimizer, epochs=50)
\end{lstlisting}

\end{document}